\documentclass[runningheads]{llncs}
\usepackage[T1]{fontenc}
\usepackage{graphicx}
\usepackage{amsmath, amssymb, amsfonts}
\usepackage{booktabs}
\usepackage{geometry}
\usepackage{enumitem}
\usepackage{algorithm}
\usepackage{algpseudocode}
\usepackage{listings}
\usepackage{xcolor}

\begin{document}
\title{Latent Heuristic Search: Continuous Optimization for Automated Algorithm Design}
\titlerunning{LSH}
%
\author{Cheikh Ahmed\inst{1}\and
Mahdi Mostajabdaveh\inst{1}\and
Zirui Zhou\inst{1}}
\authorrunning{C. Ahmed et al.}
%
\institute{Huawei Technologies Canada, Burnaby, Canada}
\maketitle              

\begin{abstract}
The integration of Large Language Models (LLMs) into evolutionary frameworks has established a new paradigm for automated heuristic discovery. Despite their promise, these methods typically search in the discrete space of program syntax, relying on stochastic sampling to navigate a highly non-convex optimization landscape. This work proposes a continuous heuristic discovery framework that shifts optimization to a learned latent manifold. We employ an encoder to map discrete programs into continuous embeddings and train a differentiable surrogate model to predict performance, enabling gradient-based search. To regularize the optimization trajectory, an invertible normalizing flow maps these embeddings to a structured Gaussian prior, where we perform gradient ascent. The resulting optimized latent vectors are projected through a learned mapper into soft prompts, which condition a frozen LLM to synthesize novel executable heuristics. We evaluate the proposed method on the Traveling Salesman Problem (TSP), the Capacitated Vehicle Routing Problem (CVRP), the Knapsack Problem (KSP), and Online Bin Packing (OBP). Empirical results demonstrate that continuous latent-space optimization achieves performance competitive with state-of-the-art discrete evolutionary baselines while offering a complementary methodological alternative for automated algorithm design. The implementation code is available at \url{https://github.com/cheikh025/LHS}.

\keywords{Heuristic discovery \and Latent program optimization \and Large language models \and Combinatorial optimization \and Normalizing flows}
\end{abstract}
\section{Introduction}
Heuristic algorithms are central to combinatorial optimization, enabling high-quality solutions for problems such as the Traveling Salesman Problem (TSP)~\cite{lkh}, the Capacitated Vehicle Routing Problem (CVRP)~\cite{VIDAL2022105643}, and online bin packing (OBP)~\cite{obp} under practical time constraints. Despite their importance, designing strong heuristics remains largely expert-driven, requiring iterative refinement, deep domain knowledge, and extensive empirical testing.

Recent work has shown that large language models (LLMs) can support algorithm discovery when embedded in a rigorous \emph{search-and-evaluate} loop. FunSearch couples LLM-based code generation with evolutionary sampling and execution-based evaluation, demonstrating improvements on mathematical construction tasks and on classic heuristic settings such as online bin packing \cite{FunSearch2023}. Building on this idea, Evolution of Heuristics (EoH) introduces a framework that co-evolves high-level natural-language ``thoughts'' describing heuristic strategies alongside executable code, with the aim of improving search efficiency and diversity \cite{eoh}. ReEvo further frames LLMs as hyper-heuristics and uses reflective feedback to guide evolutionary search, providing an explicit mechanism for steering generation toward more effective heuristic behaviors \cite{reevo}. Complementary lines of work explore alternative search operators and exploration strategies, including MCTS-based exploration \cite{mctsahd}, diversity-driven evolutionary search with Harmony Search \cite{hsevo}, and more general evolutionary coding agents for scientific and algorithmic discovery \cite{alphaevolve}. Collectively, these works highlight a key lesson: LLMs become substantially more reliable for heuristic design when paired with systematic exploration and objective evaluation, rather than one-shot generation.

However, many LLM-based automated heuristic design methods still operate primarily in the \emph{discrete space of programs}, relying on mutations, recombination, and repeated LLM sampling over code tokens. This often yields a challenging optimization landscape and can require many expensive evaluations to discover consistent improvements. In parallel, research on \emph{latent program representations} suggests that program search can be made more structured by learning continuous embeddings of code and performing optimization directly in that learned space. Approaches such as LEAPS learn embeddings and decoders that enable search over latent variables rather than explicit syntactic edits \cite{leaps}, and subsequent work explores latent program space search as a mechanism for generalization and adaptation \cite{lpn}. Closely related, \emph{program synthesis} has been explicitly framed as continuous optimization under specifications given by input/output examples or test-based error signals: NPO \cite{Liskowski2020} trains a neural program autoencoder to embed programs into a continuous space and then applies derivative-free optimization (e.g., CMA-ES) over latents, decoding candidates back to programs for execution-based scoring. In contrast, GENESYS \cite{mandal2023} avoids a learned autoencoder by directly parameterizing token choices with continuous variables and similarly optimizes these parameters with CMA-ES augmented by restart policies. These results motivate the hypothesis that automated heuristic design may benefit from shifting the primary search process from code token space to a continuous representation space that captures semantic variation among heuristics; however, our setting differs from I/O consistency synthesis in that we search for \emph{high-performing heuristic programs} under execution-based benchmarking, rather than programs that satisfy a given set of I/O examples.

In this work, we propose a framework for discovering improved heuristic algorithms by \emph{optimizing in a learned latent space of heuristic programs}. Our approach represents candidate programs as continuous embeddings and learns a surrogate ranking model that predicts heuristic performance from these representations. To regularize the optimization procedure and preserve decodability, we fit an invertible normalizing flow that maps program embeddings to a structured prior space, enabling gradient-based search in a well-conditioned space while remaining close to the manifold of plausible heuristics.

To translate optimized latent codes into executable algorithms, we introduce a decoding pathway in which candidate embeddings are mapped to soft prompts that condition an LLM-based code generator. The resulting code is validated for syntactic correctness and executed under the benchmark protocol to obtain its performance score. In contrast to token-level evolutionary search, our approach leverages a \emph{continuous, learnable search geometry} over heuristic designs while retaining strict execution-based evaluation.

\section{Problem Formulation}
\label{sec:setup}

Let $\mathcal{P}$ denote the set of syntactically valid programs, written in a fixed programming language, that implement heuristics for a target combinatorial optimization problem (e.g., TSP, OBP). For any $p \in \mathcal{P}$, we evaluate performance by executing $p$ on a benchmark set of instances $D_{\mathrm{bench}}=\{x_j\}_{j=1}^{M}$ under a standardized protocol (e.g., fixed time limits, feasibility checks, and scoring rules), which yields instance-level objective values $\mathrm{Cost}(p; x_j)\in\mathbb{R}$.

To unify notation across minimization and maximization tasks, we define
\begin{equation}
y(p) \;:=\; \frac{1}{M}\sum_{j=1}^{M} \mathrm{Cost}(p; x_j),
\qquad
s(p) \;:=\; \alpha \, y(p), \quad \alpha\in\{+1,-1\},
\end{equation}
where $\alpha=+1$ for maximization tasks and $\alpha=-1$ for minimization tasks. The mapping $p \mapsto s(p)$ is treated as a black box and is non-differentiable with respect to program tokens.

Using this evaluator, we maintain a dataset of scored programs $\mathcal{D}=\{(p_i,s_i)\}_{i=1}^{N}$, where $s_i := s(p_i)$. The goal is to discover an improved heuristic program by solving
\begin{equation}
p^\star \in \arg\max_{p \in \mathcal{P}} \; s(p),
\label{eq:program_argmax}
\end{equation}
using search in a learned continuous representation space rather than discrete, token-level edits in program space.

\section{Method}
\label{sec:method}
\begin{figure}[!ht]
  \centering
  \includegraphics[width=\linewidth]{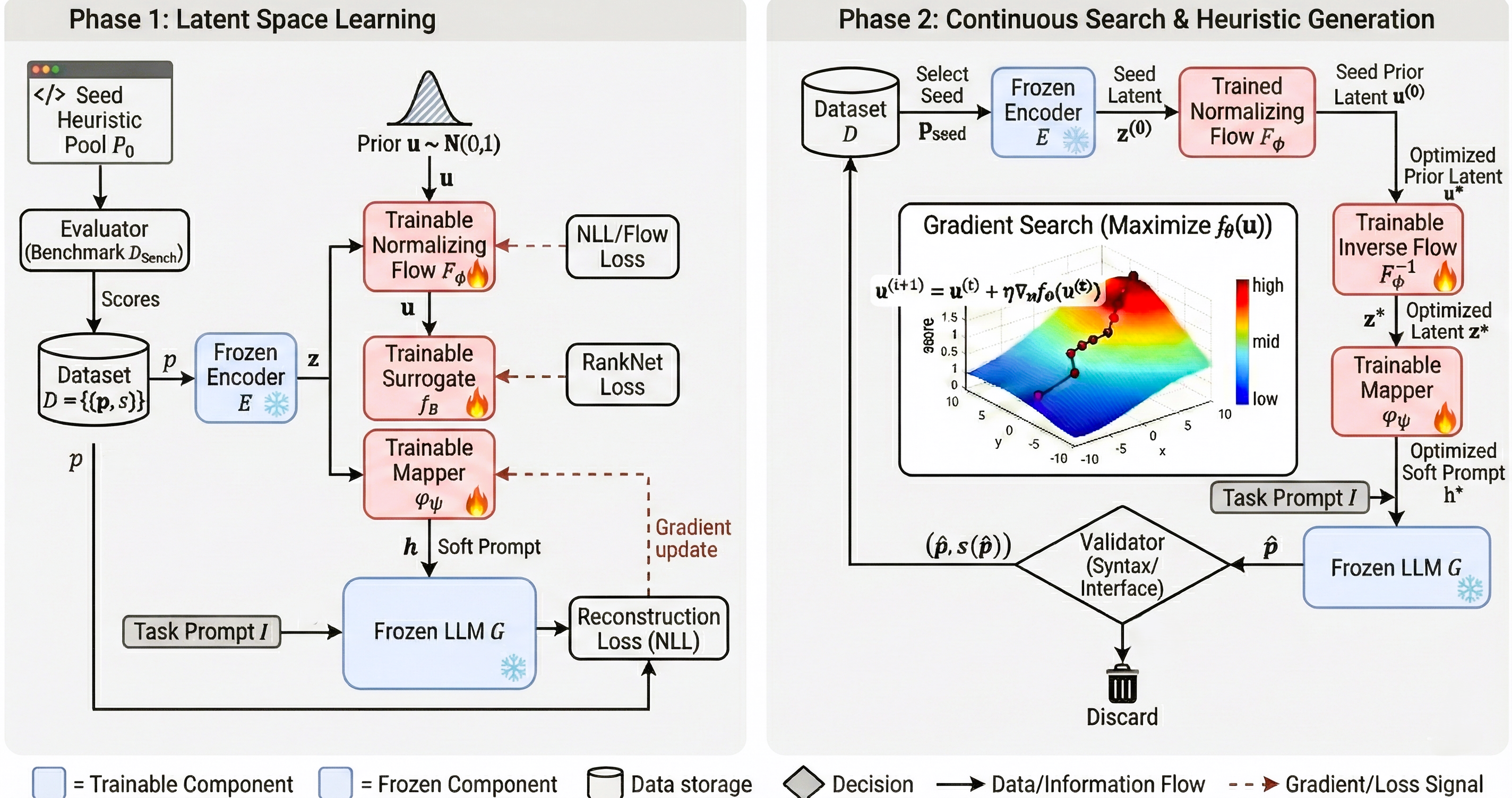}
  \label{fig:lhs_pipeline}
\end{figure}

We propose a framework for heuristic discovery via continuous optimization in a learned latent program space. Figure~\ref{fig:lhs_pipeline} provides an overview of the proposed pipeline. Specifically, we (i) encode programs into continuous representations, (ii) train a differentiable surrogate that predicts program performance from these representations, and (iii) perform gradient-based optimization to identify improved latent candidates, which are subsequently decoded into executable programs by a large language model (LLM) and evaluated under the benchmark protocol.

\subsection{Latent Heuristic Search}
\label{sec:latent_search}
To discover high-performing heuristics, we shift the search process from the discrete, non-differentiable space of code to a continuous latent manifold. This transition enables the use of continuous optimization techniques, such as gradient-based updates, in the learned representation space, The complete discovery procedure is summarized in Algorithm~\ref{alg:lhs}. Each heuristic program $p \in \mathcal{P}$ is mapped to a latent vector using a code encoder:
\begin{equation}
z \;=\; E(p), \qquad z \in \mathbb{R}^d .
\label{eq:encode}
\end{equation}
Our objective is to find programs with maximal benchmark performance score $s(p)$ (Section~\ref{sec:setup}). Since $s(p)$ is obtained by executing $p$ under the benchmark protocol, it is treated as a black-box function and does not provide gradients with respect to $z$.

To obtain a differentiable search objective, we learn a surrogate predictor $f_\theta:\mathbb{R}^d \to \mathbb{R}$. Using the dataset of evaluated programs $\mathcal{D}=\{(p_i,s_i)\}_{i=1}^N$, we encode each program as $z_i=E(p_i)$ and train $f_\theta$ such that:
\begin{equation}
f_\theta(z_i) \approx s_i .
\label{eq:surrogate_train}
\end{equation}

Given a seed latent $z^{(0)}$ (e.g., the encoding of a reference program), we generate improved latent candidates by gradient ascent on the surrogate:
\begin{equation}
z^{(t+1)} \;=\; z^{(t)} + \eta \, \nabla_z f_\theta\!\left(z^{(t)}\right),
\quad t=0,\ldots,T-1.
\label{eq:latent_ascent}
\end{equation}

After optimization, the final latent vector $z^{(T)}$ must be converted back into executable code. We project $z^{(T)}$ into the embedding space of a code-generating LLM via a learnable projection function $\phi_\omega$:
\begin{equation}
h \;=\; \phi_\omega(z^{(T)}) \in \mathbb{R}^{K \times e},
\label{eq:latent_to_prompt}
\end{equation}
which produces $K$ continuous conditioning embeddings. The sequence $h$ is used as a soft prompt to condition a frozen decoder-only LLM $\mathcal{G}$ for code generation using prefix/prompt tuning. Given a context prompt $I$ (problem description, specification, constraints), we generate a candidate program:
\begin{equation}
\hat{p} \sim \mathcal{G}\!\left(I \oplus h\right).
\label{eq:decode}
\end{equation}
where $\oplus$ denotes the concatenation of the task prompt $I$ and the soft prompt $h$. We then validate and execute $\hat{p}$ to obtain its benchmark score $s(\hat{p})$.

\subsection{Training the Latent-to-Prompt Mapper}
\label{sec:mapper_training}

The mapper network $\phi_\omega$ serves as a semantic bridge between the compact latent space and the high-dimensional input space of the decoder. It projects a program's latent vector $z$ into a sequence of $K$ continuous embedding vectors, essentially a soft prefix, that conditions the frozen Large Language Model (LLM). This approach adopts a prefix-tuning strategy~\cite{li2021prefix}, allowing us to steer the generation process without modifying the parameters of the LLM.

For each program $p_i$ in our dataset, we first compute its latent representation $z_i = E(p_i)$ using the encoder. The mapper then transforms this vector into a sequence of soft tokens:
\begin{equation}
h_i \;=\; \phi_\omega(z_i) \in \mathbb{R}^{K \times e},
\end{equation}
where $e$ denotes the embedding dimension of the decoder $\mathcal{G}$.

To ensure that $h_i$ retains the semantic information required to reconstruct the original program, we train $\psi$ using a supervised reconstruction objective. Let the program $p_i$ be tokenized as a sequence $(t_{i,1},\dots,t_{i,T_i})$, and let $I_i$ denote the fixed textual task prompt (specifying the problem definition, interface, and constraints). The decoder $\mathcal{G}$ defines an autoregressive probability distribution over the tokens:
\[
P_{\mathcal{G}}(t_{i,1:T_i}\mid I_i, h_i)
\;=\;
\prod_{k=1}^{T_i} P_{\mathcal{G}}\!\left(t_{i,k} \;\big|\; t_{i,<k}, \; I_i \oplus h_i\right).
\]
We optimize the parameters $\omega$ by minimizing the negative log-likelihood of the ground-truth program tokens under teacher forcing:
\begin{equation}
\mathcal{L}_{\mathrm{map}}(\omega)
\;=\;
-\frac{1}{N}\sum_{i=1}^{N}\frac{1}{T_i}
\sum_{k=1}^{T_i}
\log P_{\mathcal{G}}\!\left(t_{i,k} \;\big|\; t_{i,<k},\; I_i \oplus \phi_\omega(E(p_i))\right).
\label{eq:lmap}
\end{equation}
By minimizing $\mathcal{L}_{\mathrm{map}}$, the mapper learns to translate the geometric coordinates of the latent space into a conditioning signal that reliably elicits the corresponding syntax and logic from the frozen decoder $\mathcal{G}$.

\subsection{Latent optimization with a normalizing flow}
\label{sec:flow_opt}
While the encoder $E$ maps discrete heuristics to continuous vectors, performing gradient-based search directly in the raw embedding space $\mathcal{Z}$ is suboptimal. Pre-trained encoder representations are known to suffer from the anisotropy problem: their empirical distribution is typically highly non-isotropic and may contain low-density regions where the decoder becomes unreliable. Direct gradient ascent in $\mathcal{Z}$ can therefore drift toward out-of-distribution latents, yielding invalid or low-quality generated heuristics. Prior work in representation learning and controllable generation reports analogous issues, such as anisotropy, asymmetry, and non-convex support in latent spaces, and shows that learning an invertible transformation to a simple Gaussian prior space substantially stabilizes downstream optimization and control\cite{li2020sentence,gu2023controllable}.

\subsubsection{Normalizing flows.}
A \emph{normalizing flow} is an invertible mapping
$F_\varphi:\mathbb{R}^d \to \mathbb{R}^d$ that transforms samples between a complex
data distribution and a simple base distribution (typically a standard Gaussian),
while providing an exact likelihood via the change-of-variables rule.
Concretely, flows are commonly implemented as a composition of invertible layers
$F_\varphi = f_L \circ \cdots \circ f_1$ with efficient inversion and Jacobian
log-determinant computation \cite{Papamakarios,dinh2015,li2020sentence}.
Given $u = F_\varphi(z)$ and a base density $p_{\mathcal{U}}(u)$, the induced density on
$z$ is:
\begin{equation}
\log p_{\mathcal{Z}}(z)
\;=\;
\log p_{\mathcal{U}}(u)
\;+\;
\log \left|\det \frac{\partial F_\varphi(z)}{\partial z}\right|,
\qquad u = F_\varphi(z).
\label{eq:change_of_variables_flow}
\end{equation}
Invertibility is crucial in our setting because it allows us to optimize in the structured
prior space and then map the optimized point back through $F_\varphi^{-1}$ for decoding.

\subsubsection{Density Modeling}
We introduce a flow network $F_\varphi: \mathbb{R}^d \to \mathbb{R}^d$ that maps a code embedding $z$ to a latent variable $u$ in a structured search space:
\begin{equation}
u \;=\; F_\varphi(z), \quad \text{where } u \sim \mathcal{N}(0, I).
\end{equation}
By enforcing a standard Gaussian prior on $u$, the flow transform the irregular manifold of valid programs into a smooth, convex space. Optimization steps taken in $\mathcal{U}$ correspond to traversing the data manifold in $\mathcal{Z}$, preventing the search from drifting into low-density regions where the decoder behavior is undefined.

The flow is trained via maximum likelihood estimation on the dataset $\mathcal{D}$. We minimize the negative log-likelihood of the observed program embeddings under the change-of-variables formula:
\begin{equation}
\mathcal{L}_{\mathrm{flow}}(\varphi) \;=\; -\frac{1}{N} \sum_{i=1}^{N} \Big[ \log p_{\mathcal{N}}(F_\varphi(z_i)) +  \log \left|\det \frac{\partial F_\varphi(z_i)}{\partial z_i}\right| \Big],
\label{eq:flow_loss}
\end{equation}
where $p_{\mathcal{N}}$ is the standard normal density. We implement $F_\varphi$ using a sequence of invertible coupling layers (e.g., RealNVP \cite{dinh2017density}), ensuring that the Jacobian determinant is computationally efficient to compute.

\subsubsection{Search in the Prior Space}
With the flow trained, we refine the search procedure described in Section \ref{sec:latent_search}. Instead of optimizing $z$ directly, we perform gradient ascent in the prior space $\mathcal{U}$. Given a seed $z^{(0)}$, we map it to $u^{(0)} = F_\varphi(z^{(0)})$. The update rule becomes:
\begin{equation}
u^{(t+1)} \;=\; u^{(t)} + \eta \, \nabla_u f_\theta(u^{(t)}).
\end{equation}
This formulation naturally constrains the generated codes to remain "program-like," as the Gaussian prior ensures we stay within the support of the training data. After optimization, the final candidate is obtained by inverting the flow: $z^* = F_\varphi^{-1}(u^{(T)})$.

\subsection{Surrogate Training}
\label{sec:surrogate}

To guide the gradient search, we require a differentiable surrogate $f_\theta$ that approximates the black-box performance score $s(p)$. We define this surrogate on the prior space $\mathcal{U}$, such that $f_\theta: \mathcal{U} \to \mathbb{R}$.

A straightforward approach would be to train $f_\theta$ to regress the absolute score $s(p)$ via Mean Squared Error (MSE). However, we reject this approach for three critical reasons. Pointwise regression with an MSE loss attempts to match absolute scores, which can be brittle when evaluation is noisy and when the scale of $s(p)$ varies across instance sets. In contrast, the search procedure only requires a consistent ordering of candidates. Pairwise ranking losses focus precisely on this relative information, yielding a stable learning signal even when absolute score calibration is unreliable. A further advantage of ranking is that it provides a natural way to expand the supervised training signal from a small set of evaluated programs. From $N$ scored programs $\{(u_i,s_i)\}_{i=1}^{N}$, we construct preference pairs
\[
\mathcal{D}_r = \Big\{ (u_i, u_j) \;\big|\; s_i > s_j \Big\}.
\]
This yields up to $O(N^2)$ training comparisons derived from only $N$ evaluations, substantially increasing the number of supervised constraints available to fit $f_\theta$. We emphasize that these pairwise examples are not independent; accordingly, we evaluate generalization using program-level splits.

We train $f_\theta$ to minimize the RankNet loss \cite{burges2005ranknet}.
The parameters $\omega$ are optimized to maximize the likelihood of correctly predicting the superior candidate:
\begin{equation}
\mathcal{L}_{\mathrm{rank}}(\omega) = - \mathbb{E}_{(u_i, u_j) \sim \mathcal{D}_r} \Big[ \log \sigma\Big( f_\theta(u_i) - f_\theta(u_j) \Big) \Big],
\end{equation}
where $\sigma$ is the sigmoid function.

\begin{algorithm}[H]
\caption{Latent Heuristic Search (LHS)}
\label{alg:lhs}
\begin{algorithmic}[1]
\Require Pretrained encoder $E$, flow $F_\varphi$, surrogate $f_\theta$, mapper $\phi_\omega$, and frozen code LLM $\mathcal{G}$
\Require Initial seed corpus $\mathcal{P}_0$; task prompt $I$; benchmark evaluator $s(\cdot)$
\Require Budgets: outer iterations $R$, candidates per round $B$, ascent steps $T$, step size $\eta$
\State $\mathcal{D} \gets \emptyset$
\ForAll{$p \in \mathcal{P}_0$}
  \State Execute $p$ to obtain score $s(p)$
  \State $\mathcal{D} \gets \mathcal{D} \cup \{(p,s(p))\}$
\EndFor
\For{$r=1$ to $R$}
  \State Select seed programs $\{p^{(b)}\}_{b=1}^{B}$ from $\mathcal{D}$ (e.g., top-$k$)
  \For{$b=1$ to $B$}
    \State $z^{(0)} \gets E(p^{(b)})$; $u^{(0)} \gets F_\varphi(z^{(0)})$
    \For{$t=0$ to $T-1$}
      \State $u^{(t+1)} \gets u^{(t)} + \eta\,\nabla_u f_\theta\big(u^{(t)}\big)$
    \EndFor
    \State $z^* \gets F_\varphi^{-1}(u^{(T)})$; $h \gets \phi_\omega(z^*)$
    \State Sample candidate code $\hat{p} \sim \mathcal{G}(\mathrm{Concat}(I, h))$
    \If{$\hat{p}$ is valid}
      \State Execute $\hat{p}$ to obtain $s(\hat{p})$
      \State $\mathcal{D} \gets \mathcal{D} \cup \{(\hat{p}, s(\hat{p}))\}$
    \EndIf
  \EndFor
\EndFor
\State \Return best program $p^* = \arg\max_{(p,s)\in\mathcal{D}} s$
\end{algorithmic}
\end{algorithm}

\section{Experiments}
We empirically validate the proposed latent-space heuristic discovery framework across a diverse suite of combinatorial optimization tasks.
\subsection{Data Generation} \label{sec:data}

To train the LHS components, we constructed a dataset of heuristics across ten combinatorial optimization tasks. We first generated approximately 100 seed heuristics per task using GPT-5.2. We then expanded this set using a lightweight model (GPT-OSS-20B) via three augmentation strategies: syntactic rewriting, parametric tuning, and behavioral variation. After validating the generated heuristics for correctness and uniqueness, we obtained a final corpus of approximately 4,500 heuristics. This pooled dataset was used to train the global components (normalizing flow and mapper), while task-specific subsets were used to train the individual performance surrogates.

\subsection{Baselines }
\label{sec:baselines}
We compare against representative LLM-based heuristic search methods, including
\emph{FunSearch}\cite{FunSearch2023},
\emph{Evolution of Heuristics (EoH)}\cite{eoh},
\emph{ReEvo}\cite{reevo}, and
\emph{MCTS-AHD}\cite{mctsahd}.
We implement and evaluate all baselines using the \texttt{LLM4AD} platform \cite{llm4ad}.
Our experimental protocol is designed to ensure a fair comparison by isolating the effect of the search strategy from other sources of variation. Across all tasks, every method (including ours) uses the same code-generation backbone, \texttt{Qwen3-4B-Instruct-2507}, and is allocated an identical LLM budget of 100 inference calls. In addition, all methods are initialized from the same single reference program, ensuring a common starting point and comparable prior program distribution. Since several baseline methods rely on an LLM-querying procedure to generate an initial population, we adopt the same initialization strategy for our method to ensure methodological parity. For all methods, the population size is set to 10. To account for stochasticity in LLM sampling and search, each task-method experiment is repeated with five independent random seeds; objective values are reported as the mean $\pm$ standard deviation across these five seeds, and runtimes are reported as averages over the same runs. Additional implementation details are provided in the appendix. For our approach, programs are embedded using a \texttt{Qwen3-Embedding-0.6B} embedding model, which serves as the encoder $E$. All candidate heuristics are evaluated using a shared execution-based protocol in the same sandbox\cite{llm4ad}, with identical instance distributions, time limits, and feasibility checks.

\subsection{Tasks}
We formulate each task as a sequential constructive process executed by a program $p\in\mathcal{P}$. Given an instance $x$ and a partial solution state $\sigma$, the program selects an action (e.g., the next city in a tour or the bin assignment for an incoming item), which deterministically updates the state to $\sigma'$. This rollout continues until a complete feasible solution is produced or a preset resource limit is reached. The resulting solution quality is measured by the same execution-based evaluator used throughout the paper, yielding an instance-level objective $\mathrm{Cost}(p; x)$. Overall heuristic performance is then summarized by aggregating $\mathrm{Cost}(p; x)$ over a fixed validation set, consistent with the definition of $y(p)$ in Section~\ref{sec:setup}. The five-seed protocol described above is applied uniformly to the final evaluations for TSP, CVRP, OBP, and Knapsack; the task-specific settings below therefore specify only the instance distributions and time limits. For the routing tasks, search is performed on size-50 validation instances, while final evaluation also includes larger held-out instance sizes (TSP-100, TSP-200, and CVRP-100), allowing us to assess cross-size generalization beyond the search setting.
\paragraph{Traveling Salesperson Problem.}
The Traveling Salesperson Problem (TSP) requires constructing a single continuous route that visits every city in a given set exactly once before returning to the starting point. The constructive heuristic incrementally selects the next city to visit, with the sole objective of minimizing the total distance of the completed tour.
\textbf{Settings:} During the search phase, we utilize a validation set of 16 randomly generated instances of size 50 with a 30-second timeout per instance. For final evaluation (Table~\ref{tab:tsp_results}), we report performance on 100 instances of size 50, 100 instances of size 100, and 100 instances of size 200, all with a timeout of 60 seconds.
\begin{table}[H]
\centering
\setlength{\tabcolsep}{3pt}
\begin{tabular}{l c c c c c c}
\toprule
& \multicolumn{2}{c}{\textbf{TSP-50}} & \multicolumn{2}{c}{\textbf{TSP-100}} & \multicolumn{2}{c}{\textbf{TSP-200}} \\
\cmidrule(lr){2-3} \cmidrule(lr){4-5} \cmidrule(lr){6-7}
\textbf{Method} & Obj. ($\downarrow$) & Time (s) & Obj. ($\downarrow$) & Time (s) & Obj. ($\downarrow$) & Time (s) \\
\midrule
FunSearch & \(6.85 \pm 0.19\) & \(2.83\) & \(9.45 \pm 0.22\) & \(5.80\) & \(13.18 \pm 0.23\) & \(10.16\) \\
EoH & \(6.77 \pm 0.16\) & \(6.01\) & \(9.34 \pm 0.18\) & \(7.13\) & \(13.15 \pm 0.18\) & \(9.68\) \\
ReEvo & \(\mathbf{6.46 \pm 0.19}\) & \(4.10\) & \(8.98 \pm 0.31\) & \(8.42\) & \(12.58 \pm 0.36\) & \(29.61\) \\
MCTS-AHD & \(6.66 \pm 0.21\) & \(4.13\) & \(9.26 \pm 0.33\) & \(9.12\) & \(12.91 \pm 0.46\) & \(16.36\) \\
LHS (Ours) & \(6.48 \pm 0.06\) & \(3.16\) & \(\mathbf{8.93 \pm 0.12}\) & \(3.55\) & \(\mathbf{12.49 \pm 0.20}\) & \(4.86\) \\
\bottomrule
\end{tabular}
\caption{\textbf{TSP Performance.} Comparison on TSP-50, TSP-100, and TSP-200. Objective values are mean $\pm$ standard deviation over seeds; time values are seed averages. ($\downarrow$) indicates lower is better.}
\label{tab:tsp_results}
\end{table}
\paragraph{Capacitated Vehicle Routing Problem.}
The Capacitated Vehicle Routing Problem (CVRP) extends the routing challenge to a fleet of vehicles based at a central depot. The goal is to design a set of routes that serve all customers without exceeding the maximum load capacity of any single vehicle. The heuristic must ensure all demands are met while minimizing the total accumulated travel cost across the entire fleet.
\textbf{Settings:} For the search phase, we use 16 instances of size 50 with a capacity of 40 and a 30-second timeout. For final evaluation (Table~\ref{tab:cvrp_results}), we use 100 instances of size 50 and 100 instances of size 100, keeping the capacity fixed at 40 and increasing the timeout to 60 seconds.
\begin{table}[H]
\centering
\setlength{\tabcolsep}{3pt}
\begin{tabular}{l c c c c}
\toprule
& \multicolumn{2}{c}{\textbf{CVRP-50}} & \multicolumn{2}{c}{\textbf{CVRP-100}} \\
\cmidrule(lr){2-3} \cmidrule(lr){4-5}
\textbf{Method} & Obj. ($\downarrow$) & Time (s) & Obj. ($\downarrow$) & Time (s) \\
\midrule
FunSearch & \(14.03 \pm 0.14\) & \(4.82\) & \(24.34 \pm 0.11\) & \(6.79\) \\
EoH & \(13.83 \pm 0.15\) & \(7.93\) & \(24.09 \pm 0.27\) & \(4.65\) \\
ReEvo & \(13.92 \pm 0.12\) & \(3.43\) & \(24.11 \pm 0.17\) & \(6.33\) \\
MCTS-AHD & \(13.86 \pm 0.15\) & \(5.99\) & \(23.91 \pm 0.41\) & \(16.04\) \\
LHS (Ours) & \(\mathbf{13.52 \pm 0.29}\) & \(3.45\) & \(\mathbf{23.80 \pm 0.68}\) & \(3.97\) \\
\bottomrule
\end{tabular}
\caption{\textbf{CVRP Performance.} Comparison on CVRP-50 and CVRP-100. Objective values are mean $\pm$ standard deviation over five seeds; time values are seed averages. ($\downarrow$) indicates lower is better.}
\label{tab:cvrp_results}
\end{table}
\paragraph{Online Bin Packing.}
The Online Bin Packing Problem involves the sequential arrival of items, each requiring immediate and irrevocable assignment to a bin of fixed capacity. The constructive heuristic must determine whether to place each incoming item into an existing bin or open a new one, with the primary objective of minimizing the total number of bins used to accommodate the stream of items.
\textbf{Settings:} Item sizes are drawn from a Weibull distribution. During search, we validate on 5 instances of length 5k with a capacity of 100 (timeout 30s). For evaluation (Table~\ref{tab:obp_results}), we report on 10 instances of length 1k and 10 instances of length 5k (capacity 100), with a 30-second timeout.

\begin{table}[H]
\centering
\setlength{\tabcolsep}{3pt}
\begin{tabular}{l c c c c}
\toprule
& \multicolumn{2}{c}{\textbf{OBP-1k}} & \multicolumn{2}{c}{\textbf{OBP-5k}} \\
\cmidrule(lr){2-3} \cmidrule(lr){4-5}
\textbf{Method} & Obj. ($\downarrow$) & Time (s) & Obj. ($\downarrow$) & Time (s) \\
\midrule
FunSearch & \(422.50 \pm 0.00\) & \(2.74\) & \(2090.80 \pm 0.00\) & \(5.44\) \\
EoH & \(422.32 \pm 0.36\) & \(2.82\) & \(2084.92 \pm 11.76\) & \(4.21\) \\
ReEvo & \(423.22 \pm 1.59\) & \(3.19\) & \(2081.64 \pm 16.58\) & \(4.47\) \\
MCTS-AHD & \(\mathbf{421.14 \pm 2.10}\) & \(4.06\) & \(2079.92 \pm 12.58\) & \(8.67\) \\
LHS (Ours) & \(422.60 \pm 5.16\) & \(3.91\) & \(\mathbf{2066.64 \pm 18.56}\) & \(18.97\) \\
\bottomrule
\end{tabular}
\caption{\textbf{Bin Packing Performance.} Comparison on OBP-1k and OBP-5k. Objective values are mean $\pm$ standard deviation over five seeds; time values are seed averages. ($\downarrow$) indicates lower is better.}
\label{tab:obp_results}
\end{table}

\paragraph{Knapsack.}
The Knapsack Problem requires selecting a subset of items, each characterized by a specific weight and value, to maximize the total accumulated value without exceeding a strictly defined weight capacity. The heuristic operates constructively by evaluating and selecting items for inclusion, ensuring that the aggregate weight remains within the feasible limit.
\textbf{Settings:} We follow the hard instance generation protocols of Pisinger~\cite{pisinger2005}. During search, we use a mixed validation set of 32 instances (uncorrelated, weakly correlated, and strongly correlated) with capacity 100 and 50 items, utilizing a 20-second timeout. For evaluation (Table~\ref{tab:kp_results}), we test on 100 instances of each family (capacity 100, 50 items) with a 20-second timeout.
\begin{table}[H]
\centering
\setlength{\tabcolsep}{3pt}
\begin{tabular}{l c c c c c c}
\toprule
& \multicolumn{2}{c}{\textbf{Uncorrelated}} & \multicolumn{2}{c}{\textbf{Weakly Corr.}} & \multicolumn{2}{c}{\textbf{Strongly Corr.}} \\
\cmidrule(lr){2-3} \cmidrule(lr){4-5} \cmidrule(lr){6-7}
\textbf{Method} & Obj. ($\uparrow$) & Time (s) & Obj. ($\uparrow$) & Time (s) & Obj. ($\uparrow$) & Time (s) \\
\midrule
FunSearch & \(409.32 \pm 0.00\) & \(3.21\) & \(132.83 \pm 0.00\) & \(2.71\) & \(162.05 \pm 0.00\) & \(2.80\) \\
EoH & \(410.22 \pm 0.43\) & \(2.75\) & \(127.55 \pm 2.62\) & \(2.68\) & \(157.46 \pm 5.25\) & \(2.62\) \\
ReEvo & \(\mathbf{410.87 \pm 0.52}\) & \(2.92\) & \(128.16 \pm 3.10\) & \(2.82\) & \(159.36 \pm 5.01\) & \(2.89\) \\
MCTS-AHD & \(410.35 \pm 0.57\) & \(3.07\) & \(126.20 \pm 6.18\) & \(3.16\) & \(153.39 \pm 14.96\) & \(5.08\) \\
LHS (Ours) & \(409.98 \pm 0.33\) & \(2.94\) & \(\mathbf{133.27 \pm 0.22}\) & \(3.03\) & \(\mathbf{162.05 \pm 0.00}\) & \(2.96\) \\
\bottomrule
\end{tabular}
\caption{\textbf{Knapsack Performance.} Comparison by correlation type. Objective values are mean $\pm$ standard deviation over five seeds; time values are seed averages. ($\uparrow$) indicates higher is better.}
\label{tab:kp_results}
\end{table}

\paragraph{Results.}
Tables~\ref{tab:tsp_results}, \ref{tab:cvrp_results}, \ref{tab:kp_results}, and \ref{tab:obp_results} summarize the quantitative comparisons across four representative domains. On the routing tasks (Tables~\ref{tab:tsp_results} and \ref{tab:cvrp_results}), LHS achieves the best average objective on both TSP and CVRP, consistently outperforming token-space evolutionary baselines \cite{FunSearch2023,eoh,reevo} and MCTS-AHD \cite{mctsahd} without increasing the execution time of the discovered heuristics. For Knapsack (Table~\ref{tab:kp_results}), all methods are tightly clustered; ReEvo attains the best objective, while LHS remains competitive. For Online Bin Packing (Table~\ref{tab:obp_results}), MCTS-AHD achieves the best objective but with substantially slower runtime, whereas LHS matches the evolutionary baselines' objective with moderate heuristic runtime. Overall, these results demonstrate that continuous optimization over learned program representations yields state-of-the-art heuristic quality on core routing domains while maintaining practical efficiency.

\subsection{Ablation Study}
To investigate the contributions of the normalizing flow and the gradient-based optimization strategy, we conduct an ablation study comparing our proposed framework (LHS) against two variants:
\begin{itemize}
    \item \textbf{No-Flow (Optimization in $\mathcal{Z}$):} We remove the normalizing flow and perform gradient ascent directly on the raw encoder embeddings $z$. This tests the hypothesis that the flow provides a necessary regularization of the search space.
    \item \textbf{No-Grad (Interpolation):} We remove the differentiable surrogate model to test if simple geometric exploration is sufficient. Instead of gradient ascent, we generate candidates by computing the linear midpoint between pairs of top-performing programs: $u_{\text{new}} \leftarrow 0.5 u_a + 0.5 u_b$. This represents a conservative baseline that explores the center of the convex hull defined by the parent programs.
\end{itemize}

We evaluate these variants on the TSP task using the same computational budget. We report the \textit{Success Rate} (percentage of generated heuristics that compile and execute within time limits) and the \textit{Benchmark Score} (average objective cost, lower is better).

\begin{table}[h]
\centering
\setlength{\tabcolsep}{8pt}
\begin{tabular}{l c c c}
\toprule
\textbf{Method} & \textbf{Success Rate (\%)} & \textbf{Obj.} $\downarrow$ & \textbf{Runtime (s)} \\
\midrule
No-Grad (Interpolation) & \textbf{86.0} & 6.79 & 0.90 \\
No-Flow (Search in $\mathcal{Z}$) & 61.0 & 6.65 & 0.84 \\
\textbf{LHS (Search in $\mathcal{U}$)} & 74.0 & \textbf{6.61} & \textbf{0.72} \\
\bottomrule
\end{tabular}
\caption{\textbf{Ablation Analysis.} Comparison of search strategies on TSP. \textit{LHS} (Ours) achieves the best (lowest) objective cost. \textit{No-Flow} suffers from low validity due to latent space anisotropy. \textit{No-Grad} shows high stability but poor optimization capability.}
\label{tab:ablation}
\end{table}

\paragraph{Impact of the Normalizing Flow.}
Comparing LHS to the \textit{No-Flow} variant reveals the critical role of the prior space $\mathcal{U}$ in stabilizing optimization. Searching directly in the raw encoder space $\mathcal{Z}$ results in a significantly lower success rate ($61\%$ vs $74\%$). Raw encoder embeddings often suffer from anisotropy and "holes" in the manifold; gradient steps in $\mathcal{Z}$ are prone to traversing low-density regions where the decoder behavior is undefined, leading to invalid syntax or runtime errors. The normalizing flow maps these embeddings to a smooth, convex Gaussian prior, ensuring that optimization trajectories remain closer to the manifold of valid programs.

\paragraph{Gradient Search vs. Interpolation.}
The comparison with the \textit{No-Grad} baseline highlights a fundamental trade-off between optimization power and stability. Interpolation achieves the highest success rate ($86\%$) because interpolating between valid points generally keeps the candidate within the data hull. However, it yields the worst performance (6.79), as it lacks the directional signal required to discover novel regions of the landscape. Conversely, LHS leverages the surrogate gradient to actively minimize cost, achieving the best score (6.61), but with a slightly lower success rate than interpolation ($74\%$). This indicates that while the gradient is essential for quality, it introduces a risk of pushing the latent vector "off-manifold" if the optimization trajectory is too aggressive. Consequently, the learning rate $\eta$ and the number of optimization steps $T$ are critical hyperparameters that must be tuned carefully to balance the aggressiveness of the search with the validity of the generated programs.
\subsection{Visualization of Heuristics}
\label{sec:tsp_2d_viz}
\begin{figure}[t]
  \centering
  \includegraphics[width=0.8\linewidth]{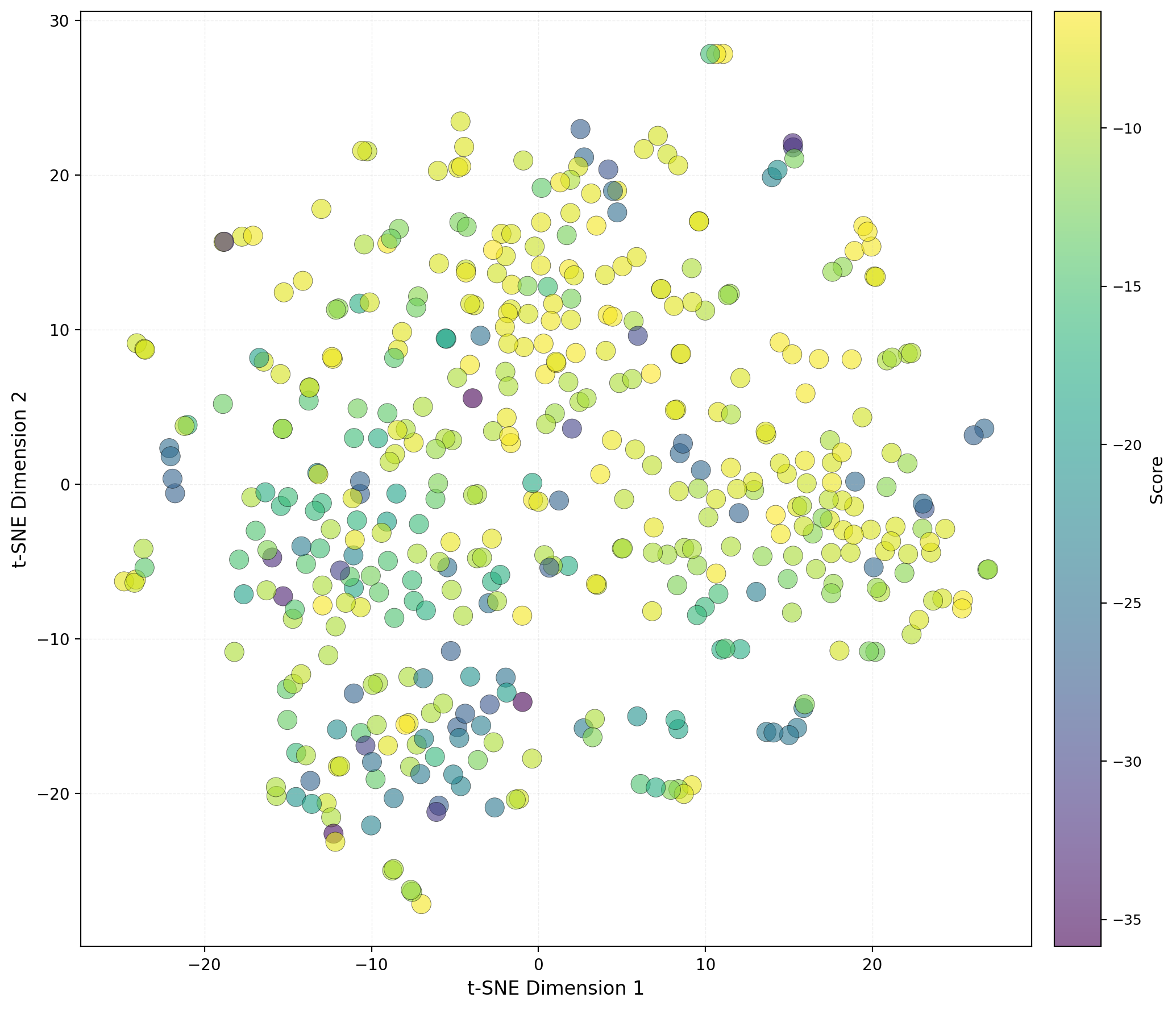}
  \caption{\textbf{Latent Space Visualization.} A 2D t-SNE projection of program embeddings for the TSP. The color gradient represents the heuristic score, showing that high-performing programs (lighter colors) cluster together rather than being randomly dispersed.}
  \label{fig:tsp_tsn}
\end{figure}

To analyze the topology of the program space, we project the TSP heuristic embeddings (learned by the encoder $E$) into two dimensions using t-SNE. Figure~\ref{fig:tsp_tsn} The t-SNE plot displays distinct clusters of heuristics. Notably, areas with concentrated yellow points represent heuristics producing shorter routes (better solutions), while darker clusters indicate less effective ones. This pattern suggests that similar heuristics tend to group together, both in terms of their characteristics and their relative performance on the TSP This behavior is consistent with the hypothesis that $E$ clusters semantically similar heuristics i.e., programs implementing related construction strategies into nearby latent neighborhoods. Consequently, continuous latent-space optimization is able to exploit this structure by moving candidates toward high-quality neighborhoods, increasing the likelihood that decoding yields strong executable heuristics.

\section{Discussion}

This work introduces a continuous alternative to token-level evolutionary search by shifting heuristic discovery into a learned latent manifold. Our results suggest that continuous optimization is a powerful complement to discrete search; it enables systematic navigation via gradient information from a differentiable surrogate while maintaining the rigor of execution-based evaluation.

Empirical results demonstrate that LHS is competitive with state-of-the-art baselines, often achieving superior objectives within a fixed evaluation budget. A key finding is the necessity of optimizing within the flow prior space $\mathcal{U}$. While searching in $\mathcal{U}$ significantly improves stability and validity compared to the raw embedding space $\mathcal{Z}$, it also highlights a critical trade-off: overly aggressive ascent steps can push candidates into regions that, while favored by the surrogate, remain brittle during decoding. This suggests that latent optimization benefits from a "trust-region" style approach to ensure candidates stay on the manifold of executable code.

The current implementation of LHS relies on a relatively small initial corpus to train the flow, mapper, and surrogate. While we expect that moving to a larger, more versatile dataset would significantly improve the stabilization of the latent manifold and the robustness of the search, scaling the data comes with a practical trade-off in terms of upfront generation and evaluation costs. This initial investment is most justifiable in scenarios involving repeated searches across related tasks or long-term algorithmic refinement. However, because the normalizing flow and mapper learn general properties of program syntax rather than task-specific logic, these components can be amortized across different optimization domains. In this multi-task regime, only the lightweight surrogate needs to be retrained for new problems, making the overall framework increasingly efficient as it is applied to broader classes of heuristics. This efficiency is critical because the efficacy of LHS is fundamentally tied to the surrogate model $f_\theta$. As optimization moves away from the initial training distribution, the surrogate may suffer from distribution shift, leading to ``false ascents'' where predicted improvements do not manifest in execution. While our ranking-based supervision provides a good signal, we expect that scaling the initial program corpus and incorporating active learning where the surrogate is iteratively retrained on newly discovered heuristics will further stabilize the search.

In conclusion, by treating programs as continuous latent vectors, we move toward a more differentiable pipeline for automated algorithm design. Future work will explore hybridizing LHS with discrete mutations to combine the global exploration of evolution with the local refinement of gradient-based search, aditionaly othrs evolution search in latent space ideas can be explored

In conclusion, by treating programs as continuous latent vectors, we move toward a more differentiable pipeline for automated algorithm design. Future work will explore hybridizing LHS with discrete evolutionary operators (e.g., mutation and recombination) to combine the global exploration of evolutionary search with the local refinement of gradient-based optimization \cite{FunSearch2023,eoh,reevo}. In addition, broader families of latent-space evolutionary strategies and population-based search methods could be incorporated\cite{hansen2023}.

%
%
%
\bibliographystyle{splncs04}
\bibliography{mybibliography}
\appendix
\newpage
\section{Implementation Details}
\label{sec:implementation}

We provide a detailed breakdown of the model architectures, training objectives, and hyperparameters used in the LHS framework. The pipeline consists of four primary components: the code encoder, the normalizing flow, the latent-to-prompt mapper, and the performance surrogate.

\subsection{Latent Space Construction (Encoder)}
To map discrete program syntax into a continuous manifold, we employ the \texttt{Qwen3-Embedding-0.6B} model. To ensure computational efficiency during gradient-based search, we leverage \textbf{Matryoshka Representation Learning}. Instead of utilizing the full high-dimensional embedding 1024, we truncate the embedding vector to a compact dimensionality of $d=128$. This 128-dimensional vector serves as the input $z$ for all subsequent modules (Flow, Mapper, and Surrogate), significantly reducing the parameter count of downstream networks without sacrificing semantic fidelity.

\subsection{Normalizing Flow Architecture}
The normalizing flow $F_\varphi: \mathbb{R}^{d} \to \mathbb{R}^{d}$ is implemented as a RealNVP (Real-valued Non-Volume Preserving) transformation. Its primary role is to map the irregular latent distribution of the encoder to a standard Gaussian prior $\mathcal{N}(0, I)$.

\begin{itemize}
    \item \textbf{Structure:} The flow consists of $4$ coupling layers. We use an alternating binary mask strategy (checkerboard pattern) where parity alternates at each layer to ensure all dimensions are transformed.
    \item \textbf{Coupling Networks:} Each coupling layer utilizes two internal multi-layer perceptrons (MLPs) to predict the scale $s$ and translation $t$ parameters. These MLPs have a single hidden layer of size $128$ with LeakyReLU activation (negative slope $0.2$).
    \item \textbf{Stability:} To prevent numerical instability during inversion, the output of the scale network is passed through a $\tanh$ activation and multiplied by a learnable scale factor (initialized to $0.8$). Additionally, \textbf{ActNorm} (Activation Normalization) layers are inserted before each coupling layer to normalize feature statistics.
    \item \textbf{Training:} The flow is trained via maximum likelihood estimation (minimizing NLL) using the AdamW optimizer with a learning rate of $1e^{-3}$ and a weight decay of $1e^{-5}$.
\end{itemize}

\subsection{Latent-to-Prompt Mapper}
To translate optimized latent vectors back into executable code, we introduce a \textbf{Low-Rank Mapper} $\phi_\omega$. Unlike standard MLP mappers which can be parameter-heavy, our architecture is designed for parameter efficiency, preventing overfitting on the relatively small heuristic datasets.

\begin{itemize}
    \item \textbf{Objective:} Map a latent vector $z \in \mathbb{R}^{128}$ to a sequence of $K=16$ continuous soft-prompt tokens compatible with the LLM decoder.
    \item \textbf{Feature Expansion:} The input $z$ is first projected linearly to a tensor of shape $(K, 128)$. Learnable positional embeddings are added to differentiate the tokens.
    \item \textbf{Self-Attention:} To capture dependencies between prompt tokens, we apply a lightweight Multi-Head Self-Attention block (2 heads, embedding dimension 128, dropout 0.1).
    \item \textbf{Shared Up-Projection:} To map the processed 128-dimensional features to the high-dimensional embedding space of the decoder (e.g., 2560 dimensions for Qwen-4B), we use a shared MLP (Pointwise Feed-Forward Network). This network projects each token individually, sharing weights across the sequence to minimize parameter count.
    \item \textbf{Total Parameters:} This low-rank design results in approximately 1.8M trainable parameters, compared to $>40$M for a standard full-rank MLP mapper.
\end{itemize}

\subsection{Performance Surrogate}
We train a task-specific differentiable surrogate $R_\theta(u)$ to predict heuristic quality from prior-space vectors $u$. The surrogate is a small MLP (2 hidden layers, width 256) with ReLU and dropout 0.1, outputting a scalar score. Supervision is provided via pairwise ranking: from evaluated programs we create preference pairs $(u_i,u_j)$ whenever $s_i>s_j$, and minimize a logistic ranking loss with temperature $\tau$:
\[
\mathcal{L}_{\text{rank}} = - \log \sigma(\tau(R_\theta(u_i)-R_\theta(u_j))).
\]

\subsection{LLM for Code Generation }
The decoding phase utilizes the \textbf{Qwen3-4B-Instruct} Large Language Model. The model is kept frozen during the training of the Mapper. We employ \textbf{Flash Attention 2} and gradient checkpointing to optimize memory usage during the soft-prompt tuning phase. During search we generate candidate code using a temperature of $0.7$ and nucleus sampling ($top\_p=0.9$).
\subsection{Prompt augmentation}
For mapper training, the prompt builder samples one of three prompt families per example: a task-specific prompt, a broader problem-class prompt, or a general coding prompt. This improves robustness to prompt phrasing at generation time and reduces overfitting to a single instruction template.

\subsection{Search Strategy and Hyperparameters}
The search process operates iteratively in the structured prior space $\mathcal{U}$. In each iteration, we select high-performing seed programs, optimize their representations via gradient ascent, and decode the results.

\begin{itemize}
    \item \textbf{Initialization (Softmax Sampling):} To balance exploitation of known high-quality heuristics with exploration of the manifold, we do not simply select the top-$k$ programs. Instead, we compute selection probabilities based on the actual benchmark scores in our database. For a pool of candidate programs, the probability of selecting program $p_i$ as a seed is given by the softmax distribution:
    \begin{equation}
        P(p_i) = \frac{\exp(s(p_i))}{\sum_j \exp(s(p_j))}
    \end{equation}
    We sample 5 seeds without replacement from the top-10 programs in the database for each search iteration.
    \item \textbf{Gradient Ascent:} We perform gradient ascent on the surrogate $f_\theta(u)$ for $T=5$ steps with a learning rate of $\eta=0.001$.
    \item \textbf{Decoding:} The optimized prior vectors $u^*$ are mapped back to latent codes $z^* = F^{-1}(u^*)$ and passed to the mapper.
\end{itemize}

\subsubsection{Search-time decoding prompts (per task)}
During latent search, candidate latents are decoded by conditioning the frozen LLM with a task-specific decoding prompt:

\begin{itemize}
    \item \textbf{TSP (constructive):} \\
    \texttt{Based on the above context, write a Python function that implements a TSP heuristic. The function should construct a tour by selecting the next city step-by-step.}
    \item \textbf{CVRP (constructive):} \\
    \texttt{Based on the above context, write a Python function that implements a CVRP heuristic. The function should construct routes by selecting the next customer step-by-step.}
    \item \textbf{Knapsack (constructive):} \\
    \texttt{Based on the above context, write a Python function that implements a Knapsack heuristic. The function should select items to include in the knapsack step-by-step.}
    \item \textbf{Online Bin Packing:} \\
    \texttt{Based on the above context, write a Python function that implements an online bin packing heuristic. The function should assign each item to a bin as it arrives.}
\end{itemize}

\section{Data Generation and Augmentation}
\label{sec:data_generation}

To train the normalizing flow and latent-to-prompt mapper, we require a diverse corpus of heuristic programs that covers a wide range of algorithmic behaviors and syntactic structures. We constructed this dataset using a two-stage process: initial seed generation followed by systematic data augmentation.

\subsection{Initial Seed Generation}
For each of the ten combinatorial optimization tasks, we generated an initial set of seed heuristics using \texttt{GPT-5.2}. The model was prompted with the task description and a template function signature to produce valid, executable Python code implementing constructive heuristics. This process yielded approximately 100 unique seed programs per task.

\subsection{Augmentation Strategy}
To expand the seed corpus into a robust training dataset, we implemented a three-tier augmentation pipeline designed to induce specific types of variation in the latent space. We utilized the following strategies:

\begin{enumerate}
    \item \textbf{Syntactic \& Structural Rewriting:} Transformations that alter the code surface form without necessarily changing the algorithm's logic. This includes variable renaming (e.g., \texttt{distance\_matrix} $\to$ \texttt{dist\_mat}), control flow modifications (e.g., swapping \texttt{for} loops with list comprehensions), and replacing mathematical operations with equivalent NumPy functions (e.g., \texttt{x**2} $\to$ \texttt{np.square(x)}).
    \item \textbf{Parameter \& Hyperparameter Tuning:} Modifications to the numerical constants within the heuristic. This involves perturbing weights in scoring functions (e.g., changing $0.8 \cdot \text{dist} + 0.2 \cdot \text{cost}$ to $0.6 \cdot \text{dist} + 0.4 \cdot \text{cost}$) and adjusting threshold values for filtering operations (e.g., top-$k$ selection).
    \item \textbf{Semantic-Preserving Behavioral Diversity:} Changes that introduce slight algorithmic variations while maintaining the core logic. Techniques include adding deterministic noise for tie-breaking, swapping aggregation methods (e.g., \texttt{np.mean} vs \texttt{np.median}), and replacing greedy selection (\texttt{argmin}) with softmin or random choice among top candidates.
\end{enumerate}

To ensure quality, all generated heuristics were validated for correct Python syntax and the presence of the required function signature. We also filtered for structural uniqueness to remove near-duplicate programs.

\subsection{Final Dataset Composition}
The final augmented dataset consists of 4,624 valid heuristic programs distributed across ten domains. Table~\ref{tab:dataset_stats} summarizes the distribution of heuristics per task.

\begin{table}[h]
\centering
\caption{Distribution of Heuristics in the Augmented Dataset}
\label{tab:dataset_stats}
\begin{tabular}{lc}
\toprule
\textbf{Task} & \textbf{Count} \\
\midrule
Admissible Set & 498 \\
Capacitated Facility Location (CFLP) & 464 \\
Capacitated Vehicle Routing (CVRP) & 466 \\
Job Shop Scheduling (JSSP) & 473 \\
Knapsack Problem & 439 \\
Online Bin Packing & 479 \\
Quadratic Assignment (QAP) & 416 \\
Set Cover & 474 \\
Traveling Salesperson (TSP) & 450 \\
VRPTW & 465 \\
\midrule
\textbf{Total} & \textbf{4,624} \\
\bottomrule
\end{tabular}
\end{table}

\section{Best Discovered Heuristics}
\label{sec:best_heuristics}

We present the best heuristics discovered by each method for the Traveling Salesperson Problem (TSP).

\subsection{Traveling Salesperson Problem (TSP)}

\subsubsection{Evolution of Heuristics (EoH)}
\begin{lstlisting}[language=Python, basicstyle=\ttfamily\scriptsize, breaklines=true, frame=single]
import numpy as np

def select_next_node(current_node: int, destination_node: int, unvisited_nodes: np.ndarray, distance_matrix: np.ndarray) -> int:
    """
    Design a novel algorithm to select the next node in each step.
    """
    if len(unvisited_nodes) == 1:
        return unvisited_nodes[0]
    
    # Compute immediate distances to unvisited nodes
    distances = distance_matrix[current_node, unvisited_nodes]
    
    # Compute centrality as before
    centrality = np.sum(distance_matrix[:, unvisited_nodes], axis=0)
    centrality = centrality / np.max(centrality) if np.max(centrality) > 0 else np.ones_like(centrality)
    
    # Temporal proximity score: inverse distance (simulated)
    proximity_score = 1 / (distances + 1e-6)
    
    # Trajectory bias: assign higher bias to nodes that are geometrically aligned with the direction of movement from current to destination
    # Compute direction vector from current to destination
    direction_vector = np.array([destination_node - current_node])  # Simplified direction as ID difference (not actual coordinates)
    
    # Simulate trajectory alignment: use node coordinates (assumed available via distance_matrix inference)  here we approximate using distance-based directional consistency
    # For novelty, we create a directional consistency score based on the angular alignment with movement direction
    # We assume a directional score based on the relative position (this is a heuristic approximation)
    # Compute pairwise differences in node IDs as a proxy for spatial direction (not accurate without coordinates)
    # Instead, we use a synthetic direction score based on distance differences
    directional_consistency = np.abs(unvisited_nodes - current_node) / (np.max(unvisited_nodes) - np.min(unvisited_nodes) + 1e-6)
    
    # Normalize directional consistency
    directional_consistency = directional_consistency / np.max(directional_consistency) if np.max(directional_consistency) > 0 else np.ones_like(directional_consistency)
    
    # Combine proximity, centrality, and trajectory bias
    # Weighted score: proximity * centrality + trajectory_bias * (1 - centrality) to encourage exploration
    bias_weight = 0.3  # adjustable parameter for exploration
    trajectory_score = directional_consistency * bias_weight
    combined_score = (proximity_score * centrality) + (trajectory_score * (1 - centrality))
    
    # Apply entropy regularization: add a small random perturbation to avoid deterministic selection
    entropy_noise = np.random.dirichlet([0.5] * len(unvisited_nodes))
    combined_score += entropy_noise * 0.05
    
    next_node_idx = np.argmax(combined_score)
    return unvisited_nodes[next_node_idx]
\end{lstlisting}
\subsubsection{FunSearch}
\begin{lstlisting}[language=Python, basicstyle=\ttfamily\scriptsize, breaklines=true, frame=single]
import numpy as np

def select_next_node(current_node: int, destination_node: int, unvisited_nodes: np.ndarray, distance_matrix: np.ndarray) -> int:
    """
    Design a novel algorithm to select the next node in each step.

    Args:
    current_node: ID of the current node.
    destination_node: ID of the destination node.
    unvisited_nodes: Array of IDs of unvisited nodes.
    distance_matrix: Distance matrix of nodes.

    Return:
    ID of the next node to visit.
    """
    if len(unvisited_nodes) == 0:
        return -1  # Should not happen in valid execution

    # Compute distances from current_node to all unvisited nodes
    distances_to_unvisited = distance_matrix[current_node, unvisited_nodes]

    # Find the index of the minimum distance
    next_node_idx = np.argmin(distances_to_unvisited)
    next_node = unvisited_nodes[next_node_idx]

    return next_node
\end{lstlisting}

\subsubsection{MCTS-AHD}
\begin{lstlisting}[language=Python, basicstyle=\ttfamily\scriptsize, breaklines=true]
import numpy as np

def select_next_node(current_node: int, destination_node: int, unvisited_nodes: np.ndarray, distance_matrix: np.ndarray) -> int:
    """
    Design a novel algorithm to select the next node in each step.
    """
    if len(unvisited_nodes) == 1:
        return unvisited_nodes[0]
    
    best_next = unvisited_nodes[0]
    best_score = float('inf')
    
    current_dists = [distance_matrix[current_node][node] for node in unvisited_nodes]
    avg_dist = np.mean(current_dists)
    
    # Dynamic connectivity threshold based on local density
    unvisited_density = len(unvisited_nodes)
    base_threshold = np.percentile(current_dists, 70)  # Focus on reachable neighbors
    dynamic_threshold = max(base_threshold, avg_dist * 1.1)
    
    # Compute shared neighbor overlap with adaptive threshold
    neighbor_counts = []
    for node in unvisited_nodes:
        shared_neighbors = 0
        for other in unvisited_nodes:
            if other != node:
                # Only count edges below dynamic threshold
                if distance_matrix[node][other] < dynamic_threshold:
                    shared_neighbors += 1
        neighbor_counts.append(shared_neighbors)
    
    # Temporal memory: penalize nodes that have been "visited" in a previous step (simulated via edge frequency)
    # Create a pseudo-frequency map of edges between unvisited nodes (based on previous step patterns)
    # Simulated via neighbor overlap recurrence: if a node has high prior connectivity, penalize it
    edge_frequency = np.zeros(len(unvisited_nodes))
    for i, node in enumerate(unvisited_nodes):
        # Count how many other unvisited nodes are within threshold distance
        freq = sum(1 for j in range(len(unvisited_nodes)) 
                   if j != i and distance_matrix[node][unvisited_nodes[j]] < dynamic_threshold)
        edge_frequency[i] = freq / len(unvisited_nodes) if len(unvisited_nodes) > 0 else 0
    
    # Destination potential field: use a non-linear decay with adaptive weight
    dest_factor = np.array([1 / (1 + distance_matrix[node][destination_node] ** 1.5) 
                            for node in unvisited_nodes])
    
    # Composite score: cost + connectivity + destination + temporal penalty
    for i, node in enumerate(unvisited_nodes):
        dist_cost = current_dists[i]
        conn_gain = neighbor_counts[i] / (unvisited_density * 1.2)  # Normalize by density
        dest_gain = dest_factor[i]
        temp_penalty = edge_frequency[i] * 0.6  # Penalize frequent patterns
        
        # Score combines components with non-linear weighting
        score = dist_cost + 0.35 * conn_gain + 0.3 * dest_gain - 0.25 * temp_penalty
        
        # Only consider nodes below average distance, with a small buffer
        if dist_cost < avg_dist * 0.95:
            if score < best_score:
                best_score = score
                best_next = node
    
    return best_next
\end{lstlisting}

\subsubsection{ReEvo}
\begin{lstlisting}[language=Python, basicstyle=\ttfamily\scriptsize, breaklines=true]
import numpy as np

def select_next_node(current_node: int, destination_node: int, unvisited_nodes: np.ndarray, distance_matrix: np.ndarray) -> int:
    """
    Design a novel algorithm to select the next node in each step.
    """
    if len(unvisited_nodes) == 1:
        return unvisited_nodes[0]
    
    # Compute direct distances to all unvisited nodes
    distances = distance_matrix[current_node, unvisited_nodes]
    
    # Simulate realistic spatial coordinates using a non-uniform, clustered layout
    # Nodes are placed in clusters with varying densities and global structure
    node_coords = np.array([
        (np.sin(i * 0.3 + np.random.rand() * 0.1) * (1 + 0.5 * np.cos(i * 0.1 + np.random.rand() * 0.2)),
         np.cos(i * 0.3 + np.random.rand() * 0.1) * (1 + 0.5 * np.sin(i * 0.1 + np.random.rand() * 0.2)))
        for i in range(len(distance_matrix))
    ])
    
    # Get coordinates of current node and unvisited nodes
    current_x, current_y = node_coords[current_node]
    unvisited_coords = node_coords[unvisited_nodes]
    
    # Compute direction vectors from current node to each unvisited node
    dx = unvisited_coords[:, 0] - current_x
    dy = unvisited_coords[:, 1] - current_y
    
    # Avoid zero direction
    direction_magnitudes = np.hypot(dx, dy)
    direction_magnitudes = np.where(direction_magnitudes == 0, 1, direction_magnitudes)
    
    # Normalize direction vectors
    direction_angles = np.arctan2(dy, dx)
    
    # Dynamic trend adaptation: update global trend using exponential moving average
    # This allows the trend to adapt over time, avoiding rigid assumptions
    # Initialize trend with a small bias to avoid instability
    alpha = 0.2  # Trend adaptation rate
    global_trend_angle = np.mean(direction_angles)
    
    # Update trend dynamically: use a weighted average of current directions
    # This makes the algorithm responsive to the evolving spatial distribution of unvisited nodes
    global_trend_angle = alpha * np.mean(direction_angles) + (1 - alpha) * global_trend_angle
    
    # Compute directional deviation from the global trend
    angle_diffs = direction_angles - global_trend_angle
    # Use circular distance to avoid discontinuities
    angle_diffs = np.abs(angle_diffs)
    angle_diffs = np.minimum(angle_diffs, 2 * np.pi - angle_diffs)
    
    # Directional alignment score: higher alignment = higher score
    directional_score = 1 / (1 + angle_diffs)
    
    # Weighted distance: prefer paths aligned with the dynamic spatial trend
    weighted_distances = distances / (1 + directional_score)  # Lower values when aligned
    
    # Spatial clustering: favor nodes that are part of dense clusters (high local density)
    # Compute pairwise distances between unvisited nodes to infer cluster structure
    unvisited_distance_matrix = distance_matrix[unvisited_nodes[:, None], unvisited_nodes]
    # Compute local density: number of neighbors within a small radius
    # Use a threshold to identify clusters
    radius_threshold = 0.5
    # Compute distance to each unvisited node's neighbors
    distances_to_neighbors = unvisited_distance_matrix
    # Thresholded density: count how many neighbors are within radius
    neighbor_count = np.sum(distances_to_neighbors <= radius_threshold, axis=1)
    # Normalize density to range [0,1] and add as a clustering bonus
    clustering_bonus = (neighbor_count + 1) / (len(unvisited_nodes) + 1)
    
    # Weighted distance includes clustering preference: prefer nodes in dense regions
    weighted_distances *= clustering_bonus
    
    # Destination avoidance: avoid visiting nodes too close to destination prematurely
    if destination_node in unvisited_nodes:
        dest_dist = distance_matrix[current_node, destination_node]
        dest_penalty = np.zeros(len(unvisited_nodes))
        for idx, node in enumerate(unvisited_nodes):
            if node != destination_node:
                # Penalize nodes that are close to destination
                dist_to_dest = distance_matrix[current_node, node]
                # Use a sigmoid-like penalty that increases as distance to destination decreases
                # This encourages spreading out to avoid converging too early
                dist_diff = abs(dist_to_dest - dest_dist)
                penalty = np.exp(-0.5 * (dist_diff / 1.5)**2)
                dest_penalty[idx] = penalty
            else:
                dest_penalty[idx] = 1.0
        weighted_distances += dest_penalty
    
    # Final selection: pick the node with the minimum weighted distance
    next_node_idx = np.argmin(weighted_distances)
    next_node = unvisited_nodes[next_node_idx]
    
    return next_node
\end{lstlisting}

\subsubsection{LHS (Ours)}
\begin{lstlisting}[language=Python, basicstyle=\ttfamily\scriptsize, breaklines=true]
import numpy as np

def select_next_node(current_node: int, destination_node: int, unvisited_nodes: np.ndarray, distance_matrix: np.ndarray) -> int:
    if unvisited_nodes.size == 0:
        return int(destination_node)

    cand = unvisited_nodes.astype(int)
    r = distance_matrix[current_node, cand]
    d = distance_matrix[np.ix_(cand, cand)].copy()

    # Determine the number of unvisited nodes
    n = int(np.sum(unvisited_nodes == cand))

    # Normalize the distances using a soft-min operation
    d = np.exp(-d / (1e-6 + np.max(d) + 1e-6))

    # Calculate a weighted score based on the distances and the number of remaining nodes
    score = (r + (d.sum(axis=1) * (1.0 / (n + 1e-6)))).clip(0, None)

    # Use deterministic noise for tie-breaking
    noise = np.arange(len(score)) * 1e-9
    score += noise

    return int(cand[int(np.argmin(score))])
\end{lstlisting}

\subsection{Capacitated Vehicle Routing Problem (CVRP)}

\subsubsection{Evolution of Heuristics (EoH)}
\begin{lstlisting}[language=Python, basicstyle=\ttfamily\scriptsize, breaklines=true, frame=single]
import numpy as np

def select_next_node(current_node: int, depot: int, unvisited_nodes: np.ndarray, rest_capacity: np.ndarray, demands: np.ndarray, distance_matrix: np.ndarray) -> int:
    """Design a novel algorithm to select the next node in each step."""
    if len(unvisited_nodes) == 0:
        return -1
    
    distances = distance_matrix[current_node][unvisited_nodes]
    demands_arr = demands[unvisited_nodes]
    
    # Compute immediate feasibility
    feasible_mask = demands_arr <= rest_capacity
    
    # Compute entropy of demand distribution (uncertainty in remaining demand)
    # If all unvisited nodes have similar demands, the system is in high uncertainty
    # We penalize nodes that lead to high entropy in future demand distribution
    from scipy.stats import entropy
    from numpy import log
    
    # Group unvisited nodes by demand level to simulate clustering
    demand_bins = np.histogram(demands_arr, bins=5, range=(0, np.max(demands_arr) + 1))[1]
    demand_groups = np.digitize(demands_arr, demand_bins)
    
    # Compute group-wise entropy of demand distribution after selecting a node
    # Entropy reduction = -sum(p_i * log(p_i)) after node selection
    # We aim to reduce uncertainty by selecting nodes that help "compress" the demand distribution
    
    # Compute the prior entropy of unvisited nodes
    prior_prob = np.bincount(demand_groups, minlength=5) / len(unvisited_nodes)
    prior_entropy = -np.sum(prior_prob * np.log(prior_prob + 1e-6)) if np.any(prior_prob > 0) else 0
    
    # For each node, compute the new entropy after selection
    node_entropy_change = []
    for i, node_id in enumerate(unvisited_nodes):
        if not feasible_mask[i]:
            continue
            
        # Simulate removal of this node from the unvisited set
        new_demands = demands_arr.copy()
        new_demands[i] = 0  # temporarily remove node's demand
        new_demand_groups = np.digitize(new_demands, demand_bins)
        new_prob = np.bincount(new_demand_groups, minlength=5) / (len(unvisited_nodes) - 1)
        new_entropy = -np.sum(new_prob * np.log(new_prob + 1e-6)) if np.any(new_prob > 0) else 0
        
        # Entropy reduction = prior_entropy - new_entropy
        entropy_reduction = prior_entropy - new_entropy
        node_entropy_change.append(entropy_reduction)
    
    if len(node_entropy_change) == 0:
        # Fallback to proximity and capacity
        feasible_mask = feasible_mask & (distances <= np.max(distances) * 0.9)
        if not np.any(feasible_mask):
            return unvisited_nodes[np.argmin(distances)]
        
        proximity_score = 1 / (distances[feasible_mask] + 1e-6)
        utilization_score = np.clip(demands_arr[feasible_mask] / (rest_capacity + 1e-6), 0.2, 0.6)
        final_score = proximity_score * 0.5 + utilization_score * 0.3
        return unvisited_nodes[np.argmax(final_score)]
    
    # Combine entropy gain with proximity and capacity
    proximity_score = 1 / (distances[feasible_mask] + 1e-6)
    utilization_score = np.clip(demands_arr[feasible_mask] / (rest_capacity + 1e-6), 0.2, 0.6)
    entropy_score = np.array(node_entropy_change)[feasible_mask]
    
    final_score = proximity_score * 0.4 + utilization_score * 0.3 + entropy_score * 0.3
    
    best_idx = np.argmax(final_score)
    return unvisited_nodes[feasible_mask][best_idx]
\end{lstlisting}

\subsubsection{FunSearch}
\begin{lstlisting}[language=Python, basicstyle=\ttfamily\scriptsize, breaklines=true, frame=single]
import numpy as np

def select_next_node(current_node: int, depot: int, unvisited_nodes: np.ndarray, rest_capacity: np.ndarray, demands: np.ndarray, distance_matrix: np.ndarray) -> int:
    """Design a novel algorithm to select the next node in each step."""
    best_score = -1
    next_node = -1

    # Precompute distances and demands for unvisited nodes
    unvisited_distances = distance_matrix[current_node][unvisited_nodes]
    unvisited_demands = demands[unvisited_nodes]

    # Thresholds for flexibility and diversity
    FLEX_THRESHOLD = 0.1  # Future capacity < 10% of demand -> risky
    DEPOT_DISTANCE = distance_matrix[current_node][depot]
    DEMAND_THRESHOLD = 0.5 * np.max(unvisited_demands) if unvisited_demands.size > 0 else 0

    for i, node in enumerate(unvisited_nodes):
        demand = unvisited_demands[i]
        distance = unvisited_distances[i]

        # Skip if demand exceeds capacity
        if demand > rest_capacity:
            continue

        # Avoid division by zero
        if distance == 0:
            # Only depot has zero distance; we skip non-depot zero-distance nodes
            if node == depot:
                score = 0  # Safe, no cost
            else:
                score = float('inf')
            base_score = score
        else:
            # Base score: demand / distance (higher is better)
            base_score = demand / distance

            # --- 1. Future Capacity Flexibility Penalty ---
            future_capacity = rest_capacity - demand
            if future_capacity < FLEX_THRESHOLD * demand:
                # This visit leaves very little buffer -> risky for future small nodes
                base_score *= 2.5  # Strong penalty

            # --- 2. Proximity to Depot: Bonus for being closer than going directly to depot ---
            if distance < DEPOT_DISTANCE:
                base_score *= 1.15

            # --- 3. Diversity Bonus: Favor distant, dissimilar nodes ---
            # Compute similarity to other nodes in the neighborhood (simple heuristic)
            # Nodes with similar demand and short distance -> cluster -> penalize
            # We compute average demand of nearby nodes (within a small distance threshold)
            distance_threshold = 0.7 * DEPOT_DISTANCE
            similar_nodes = []
            for j, other_node in enumerate(unvisited_nodes):
                if other_node != node:
                    d = distance_matrix[current_node][other_node]
                    if d < distance_threshold and unvisited_demands[j] > 0.5 * demand:
                        similar_nodes.append(other_node)

            # If this node is similar to many others in a tight cluster -> penalize
            if len(similar_nodes) > 1:
                # Penalize if it's part of a small cluster
                base_score *= 0.85  # Reduce score for clustering

            # --- 4. Bonus for small-demand, nearby nodes (local refinement) ---
            if demand < DEMAND_THRESHOLD and distance < DEPOT_DISTANCE * 0.6:
                base_score *= 1.2  # Encourage visiting small, nearby nodes

            # --- 5. Global diversity bonus: if node is far from depot (good for exploration) ---
            dep_dist = distance_matrix[node][depot]  # Distance from node to depot
            if dep_dist > 0.8 * DEPOT_DISTANCE:
                base_score *= 1.2  # Encourages visiting nodes far from depot

    # Final selection
    if next_node == -1:
        # Fallback: pick closest unvisited node
        closest_dist = np.inf
        closest_node = -1
        for node in unvisited_nodes:
            d = distance_matrix[current_node][node]
            if d < closest_dist:
                closest_dist = d
                closest_node = node
        return closest_node if closest_node != -1 else unvisited_nodes[0]

    return next_node
\end{lstlisting}

\subsubsection{MCTS-AHD}
\begin{lstlisting}[language=Python, basicstyle=\ttfamily\scriptsize, breaklines=true, frame=single]
import numpy as np

def select_next_node(current_node: int, depot: int, unvisited_nodes: np.ndarray, rest_capacity: np.ndarray, demands: np.ndarray, distance_matrix: np.ndarray) -> int:
    """Design a novel algorithm to select the next node in each step."""
    if len(unvisited_nodes) == 0:
        return -1
    
    n = len(unvisited_nodes)
    time_to_serve = np.zeros(n)
    capacity_gap_penalty = np.zeros(n)
    
    for i, node in enumerate(unvisited_nodes):
        dist = distance_matrix[current_node][node]
        demand_util = demands[node] / (rest_capacity + 1e-6)
        
        # Demand urgency: higher demand leads to higher urgency
        urgency = demands[node] * 1.5
        
        # Capacity gap penalty: penalize nodes where demand exceeds median demand of unvisited nodes
        demand_list = [demands[n] for n in unvisited_nodes]
        median_demand = np.median(demand_list)
        capacity_gap = max(0, demands[node] - median_demand)
        capacity_gap_penalty[i] = capacity_gap * 2.0
        
        # Time-to-serve scaled by urgency and capacity gap penalty
        time_to_serve[i] = dist * (1 + urgency + capacity_gap_penalty[i])
    
    # Dynamic threshold: 75th percentile to allow more flexibility while maintaining balance
    threshold = np.percentile(time_to_serve, 75)
    
    # Filter nodes below threshold (efficient and capacity-aware)
    eligible_nodes = unvisited_nodes[time_to_serve < threshold]
    
    if len(eligible_nodes) > 0:
        # Among eligible, pick the one with minimum distance
        best_idx = np.argmin([distance_matrix[current_node][node] for node in eligible_nodes])
        return eligible_nodes[best_idx]
    
    # Fallback to nearest neighbor
    best_idx = np.argmin([distance_matrix[current_node][node] for node in unvisited_nodes])
    return unvisited_nodes[best_idx]
\end{lstlisting}

\subsubsection{ReEvo}
\begin{lstlisting}[language=Python, basicstyle=\ttfamily\scriptsize, breaklines=true, frame=single]
import numpy as np

def select_next_node(current_node: int, depot: int, unvisited_nodes: np.ndarray, rest_capacity: np.ndarray, demands: np.ndarray, distance_matrix: np.ndarray) -> int:
    """Design a novel algorithm to select the next node in each step."""
    best_score = -1
    next_node = -1

    for node in unvisited_nodes:
        demand = demands[node]
        distance = distance_matrix[current_node][node]

        if distance == 0:
            continue  # Skip zero distance (same node or unreachable)
        
        if demand > rest_capacity:
            continue  # Skip nodes that exceed remaining capacity

        # Compute capacity utilization to reward efficient load usage
        capacity_utilization = (rest_capacity - demand) / rest_capacity if rest_capacity > 0 else 0
        
        # Prioritize high demand relative to distance, with amplification for underutilized capacity
        # Efficiency score emphasizes high-demand, low-distance nodes
        efficiency_score = demand / (distance + 1e-8)
        
        # Capacity bias: amplify for underutilized capacity and partial load efficiency
        # 1.5 multiplier for underutilized capacity encourages filling gaps
        # 0.3 multiplier for remaining capacity incentivizes balanced load distribution
        capacity_bias = 1.5 * capacity_utilization + 0.3 * (rest_capacity / (demand + 1e-6))
        
        # Final score: balance between demand efficiency and capacity utilization
        score = efficiency_score * (1.0 + capacity_bias)

        if score > best_score:
            best_score = score
            next_node = node

    # Fallback: if no valid node found, return the first unvisited node
    return next_node if next_node != -1 else unvisited_nodes[0] if len(unvisited_nodes) > 0 else depot
\end{lstlisting}

\subsubsection{LHS (Ours)}
\begin{lstlisting}[language=Python, basicstyle=\ttfamily\scriptsize, breaklines=true, frame=single]
import numpy as np

def select_next_node(current_node: int, depot: int, unvisited_nodes: np.ndarray, rest_capacity: float, demands: np.ndarray, distance_matrix: np.ndarray) -> int:
    if len(unvisited_nodes) == 0:
        return depot
    
    cap = float(rest_capacity)
    feasible = np.array([n for n in unvisited_nodes if demands[n] <= cap], dtype=int)
    if len(feasible) == 0:
        return depot
    
    d_cur = distance_matrix[current_node, feasible]
    d_dep = distance_matrix[depot, feasible]
    
    d_cur_adj = d_cur - d_dep
    d_cur_adj = np.clip(d_cur_adj, -np.inf, 1e12)
    
    d_cur_adj = np.clip(d_cur_adj, -np.inf, 1e12)
    return int(feasible[np.argmin(d_cur_adj)])
\end{lstlisting}

\subsection{Knapsack Problem (KSP)}

\subsubsection{Evolution of Heuristics (EoH)}
\begin{lstlisting}[language=Python, basicstyle=\ttfamily\scriptsize, breaklines=true, frame=single]
def select_next_item(remaining_capacity: int, remaining_items: List[Tuple[int, int, int]]) -> Tuple[int, int, int] | None:
    """
    Select the item with the highest value-to-weight ratio that fits in the remaining capacity.

    Args:
        remaining_capacity: The remaining capacity of the knapsack.
        remaining_items: List of tuples containing (weight, value, index) of remaining items.

    Returns:
        The selected item as a tuple (weight, value, index), or None if no item fits.
    """
    if not remaining_items:
        return None
    
    best_item = None
    max_ratio = -1
    
    for weight, value, index in remaining_items:
        if weight <= remaining_capacity:
            ratio = value / weight
            if ratio > max_ratio:
                max_ratio = ratio
                best_item = (weight, value, index)
    
    return best_item if best_item else None
\end{lstlisting}

\subsubsection{FunSearch}
\begin{lstlisting}[language=Python, basicstyle=\ttfamily\scriptsize, breaklines=true, frame=single]
def select_next_item(remaining_capacity: int, remaining_items: List[Tuple[int, int, int]]) -> Tuple[int, int, int] | None:
    """
    Select the item with the highest value-to-weight ratio that fits in the remaining capacity.

    Args:
        remaining_capacity: The remaining capacity of the knapsack.
        remaining_items: List of tuples containing (weight, value, index) of remaining items.

    Returns:
        The selected item as a tuple (weight, value, index), or None if no item fits.
    """
    best_item = None
    best_ratio = -1  # Initialize with a negative value to ensure any item will have a higher ratio

    for item in remaining_items:
        weight, value, index = item
        if weight <= remaining_capacity:
            ratio = value / weight  # Calculate value-to-weight ratio
            if ratio > best_ratio:
                best_ratio = ratio
                best_item = item

    return best_item
\end{lstlisting}

\subsubsection{MCTS-AHD}
\begin{lstlisting}[language=Python, basicstyle=\ttfamily\scriptsize, breaklines=true, frame=single]
def select_next_item(remaining_capacity: int, remaining_items: List[Tuple[int, int, int]]) -> Tuple[int, int, int] | None:
    """
    Select the item with the highest value-to-weight ratio that fits in the remaining capacity.

    Args:
        remaining_capacity: The remaining capacity of the knapsack.
        remaining_items: List of tuples containing (weight, value, index) of remaining items.

    Returns:
        The selected item as a tuple (weight, value, index), or None if no item fits.
    """
    if not remaining_items:
        return None
    
    best_item = None
    best_ratio = -1
    adjustment_factor = 1.0  # Dynamic factor that increases as capacity decreases
    
    # Adjust factor based on remaining capacity: more aggressive in tight spaces
    if remaining_capacity <= 10:
        adjustment_factor = 1.5
    elif remaining_capacity <= 20:
        adjustment_factor = 1.2
    
    # Apply adaptive ratio calculation with a smoothing term
    for weight, value, index in remaining_items:
        if weight <= remaining_capacity:
            # Modified ratio: value/weight * adjustment factor + smoothing term based on item consistency
            ratio = (value / weight) * adjustment_factor + (value / 100)  # Small consistency bonus
            if ratio > best_ratio:
                best_ratio = ratio
                best_item = (weight, value, index)
    
    return best_item
\end{lstlisting}

\subsubsection{ReEvo}
\begin{lstlisting}[language=Python, basicstyle=\ttfamily\scriptsize, breaklines=true, frame=single]
def select_next_item(remaining_capacity: int, remaining_items: List[Tuple[int, int, int]]) -> Tuple[int, int, int] | None:
    """
    Select the item with the highest value-to-weight ratio that fits in the remaining capacity.

    Args:
        remaining_capacity: The remaining capacity of the knapsack.
        remaining_items: List of tuples containing (weight, value, index) of remaining items.

    Returns:
        The selected item as a tuple (weight, value, index), or None if no item fits.
    """
    if not remaining_items:
        return None
    
    # Track frequency of each item index to promote diversity
    item_freq = {}
    for item in remaining_items:
        index = item[2]
        item_freq[index] = item_freq.get(index, 0) + 1
    
    # Identify underrepresented items (frequency < 2) to encourage diversity
    underrepresented = [item for item in remaining_items if item_freq[item[2]] < 2]
    
    # Phase detection: early, mid, late based on capacity and item count
    total_items = len(remaining_items)
    capacity_threshold = remaining_capacity
    
    # Early phase: high capacity, many items -> prioritize high ratio and value
    if total_items > 5 and capacity_threshold > 30:
        best_item = None
        best_ratio = -1
        best_value = -1
        
        for item in remaining_items:
            weight, value, index = item
            if weight <= remaining_capacity:
                ratio = value / weight
                if (ratio > best_ratio or 
                    (ratio == best_ratio and value > best_value)):
                    best_ratio = ratio
                    best_value = value
                    best_item = item
        return best_item
    
    # Mid-phase: moderate capacity -> balance value, ratio, and diversity
    elif total_items > 3 and capacity_threshold > 10:
        candidates = []
        for item in remaining_items:
            weight, value, index = item
            if weight <= remaining_capacity:
                # Dynamic penalty for high-weight items (e.g., > 0.6 capacity)
                weight_penalty = 0.0
                if weight > remaining_capacity * 0.6:
                    weight_penalty = 1.5
                
                # Diversity boost for underrepresented items
                diversity_factor = 1.0 if item_freq.get(index, 0) < 2 else 0.7
                
                # Adjusted score: value, ratio, and diversity
                ratio = value / weight
                score = (ratio * value * diversity_factor) - weight_penalty
                
                candidates.append((score, weight, value, index))
        
        if not candidates:
            return None
            
        candidates.sort(key=lambda x: x[0], reverse=True)
        selected = candidates[0]
        return (selected[1], selected[2], selected[3])
    
    # Late phase: low capacity -> prioritize stability, diversity, and small items
    else:
        candidates = []
        for item in remaining_items:
            weight, value, index = item
            if weight <= remaining_capacity:
                # Strong penalty for large items; favor small, diverse, high-value items
                weight_penalty = 0.0
                if weight > remaining_capacity * 0.4:
                    weight_penalty = 2.0
                
                # Strong diversity bonus for underrepresented items
                diversity_factor = 1.5 if item_freq.get(index, 0) < 2 else 0.5
                
                # Score: value + diversity, penalize large weights
                ratio = value / weight
                score = (ratio * value * diversity_factor) - weight_penalty
                
                candidates.append((score, weight, value, index))
        
        if not candidates:
            return None
            
        candidates.sort(key=lambda x: x[0], reverse=True)
        selected = candidates[0]
        return (selected[1], selected[2], selected[3])
\end{lstlisting}

\subsubsection{LHS (Ours)}
\begin{lstlisting}[language=Python, basicstyle=\ttfamily\scriptsize, breaklines=true, frame=single]
import numpy as np
from typing import List, Tuple, Optional

def select_next_item(remaining_capacity: int, remaining_items: List[Tuple[int, int, int]]) -> Optional[Tuple[int, int, int]]:
    fit = [(w, v, idx) for (w, v, idx) in remaining_items if w <= remaining_capacity]
    if not fit:
        return None

    for w, v, idx in fit:
        if w == 0:
            return (w, v, idx)

    # Compute densities of all remaining items
    densities = [float(v) / max(1e-12, float(w)) for (w, v, idx) in fit]
    median = np.median(densities)
    mean  = np.mean(densities)

    # Choose the best item based on a deterministic scoring function
    best_score = -np.inf
    best_item = None
    for w, v, idx in fit:
        if w == 0:
            score = v
        else:
            score = (v / max(1e-12, float(w))) * (1.0 + 0.5 * (w / remaining_capacity))
        if score > best_score:
            best_score = score
            best_item = (w, v, idx)

    return best_item
\end{lstlisting}

\subsection{Online Bin Packing (OBP)}

\subsubsection{Evolution of Heuristics (EoH)}
\begin{lstlisting}[language=Python, basicstyle=\ttfamily\scriptsize, breaklines=true, frame=single]
import numpy as np

def priority(item: float, bins: np.ndarray) -> np.ndarray:
    """Returns priority with which we want to add item to each bin.
    Args:
        item: Size of item to be added to the bin.
        bins: Array of capacities for each bin.
    Return:
        Array of same size as bins with priority score of each bin.
    """
    return 1 / (bins + 1e-8) / (bins / (bins + item))
\end{lstlisting}

\subsubsection{FunSearch}
\begin{lstlisting}[language=Python, basicstyle=\ttfamily\scriptsize, breaklines=true, frame=single]
import numpy as np

def priority(item: float, bins: np.ndarray) -> np.ndarray:
    """Returns priority with which we want to add item to each bin.
    Args:
        item: Size of item to be added to the bin.
        bins: Array of capacities for each bin.
    Return:
        Array of same size as bins with priority score of each bin.
    """
    return item - bins
\end{lstlisting}

\subsubsection{MCTS-AHD}
\begin{lstlisting}[language=Python, basicstyle=\ttfamily\scriptsize, breaklines=true, frame=single]
import numpy as np

def priority(item: float, bins: np.ndarray) -> np.ndarray:
    """Returns priority with which we want to add item to each bin.
    Args:
        item: Size of item to be added to the bin.
        bins: Array of capacities for each bin.
    Return:
        Array of same size as bins with priority score of each bin.
    """
    if item <= 0:
        return np.ones_like(bins) * 1e-8
    
    scaled_bins = bins / item
    
    # Dynamic sharpness of Lorentzian peak using sigmoid based on item size
    max_ratio = 1.4
    min_ratio = 0.7
    base_width = 0.4
    sensitivity_factor = 1.0 + 0.5 * np.log1p(1.0 / (1.0 + np.exp(-2 * (item - 1.0))))
    width = base_width * sensitivity_factor
    
    # Lorentzian kernel with adaptive width
    lorentzian_kernel = 1.0 / (1.0 + ((scaled_bins - 1.0) / width) ** 2)
    
    # Fixed threshold to suppress extreme deviations
    threshold_factor = 0.8
    lorentzian_kernel = lorentzian_kernel * threshold_factor
    
    # Cosine modulation to break symmetry and add oscillation
    phase_shift = np.pi * (scaled_bins - 0.5)
    cosine_modulation = np.cos(phase_shift + np.pi / 6) * 0.15
    modulated = lorentzian_kernel + cosine_modulation
    
    # Symmetric entropy regularization based on distance from median ratio
    median_ratio = np.median(scaled_bins)
    entropy_term = np.clip(1.0 / (1.0 + 0.4 * (np.abs(scaled_bins - median_ratio) / 0.4) ** 2), 0.4, 1.0)
    final = modulated * entropy_term
    
    # Clamp to ensure positive values and minimal baseline
    return np.clip(final + 1e-8, 1e-8, 1.0)
\end{lstlisting}

\subsubsection{ReEvo}
\begin{lstlisting}[language=Python, basicstyle=\ttfamily\scriptsize, breaklines=true, frame=single]
import numpy as np

def priority(item: float, bins: np.ndarray) -> np.ndarray:
    """Returns priority with which we want to add item to each bin.
    Args:
        item: Size of item to be added to the bin.
        bins: Array of capacities for each bin.
    Return:
        Array of same size as bins with priority score of each bin.
    """
    residual_capacity = bins - item
    feasible_residues = np.clip(residual_capacity, 0, None)
    
    if np.any(feasible_residues > 0):
        # Use inverse of residual capacity to prioritize bins with smaller remaining space
        # This ensures bins with more space get lower priority (less preferred) and vice versa
        # Normalize to balance priorities and avoid bias from large gaps
        inverse_residues = 1.0 / (feasible_residues + 1e-8)
        normalized_priority = inverse_residues / inverse_residues.sum()
        priority = normalized_priority
    else:
        priority = np.zeros_like(bins)
    
    return priority
\end{lstlisting}

\subsubsection{LHS (Ours)}
\begin{lstlisting}[language=Python, basicstyle=\ttfamily\scriptsize, breaklines=true, frame=single]
import numpy as np

def priority(item: float, bins: np.ndarray) -> np.ndarray:
    a = bins.copy().astype(float)
    a[a < item] = -np.inf
    return -np.clip(a, 0, None)
\end{lstlisting}

\end{document}